\documentclass[11pt,a4paper]{article}
\usepackage{times,latexsym}
\usepackage{url}
\usepackage[T1]{fontenc}
\usepackage{authblk} 
\usepackage{bbm}
\makeatletter
\newcommand\email[2][]%
   {\newaffiltrue\let\AB@blk@and\AB@pand
      \if\relax#1\relax\def\AB@note{\AB@thenote}\else\def\AB@note{\relax}%
        \setcounter{Maxaffil}{0}\fi
      \begingroup
        \let\protect\@unexpandable@protect
        \def\thanks{\protect\thanks}\def\footnote{\protect\footnote}%
        \@temptokena=\expandafter{\AB@authors}%
        {\def\\{\protect\\\protect\Affilfont}\xdef\AB@temp{#2}}%
         \xdef\AB@authors{\the\@temptokena\AB@las\AB@au@str
         \protect\\[\affilsep]\protect\Affilfont\AB@temp}%
         \gdef\AB@las{}\gdef\AB@au@str{}%
        {\def\\{, \ignorespaces}\xdef\AB@temp{#2}}%
        \@temptokena=\expandafter{\AB@affillist}%
        \xdef\AB@affillist{\the\@temptokena \AB@affilsep
          \AB@affilnote{}\protect\Affilfont\AB@temp}%
      \endgroup
       \let\AB@affilsep\AB@affilsepx
}
\makeatother
\usepackage{graphicx}
\usepackage{framed}

\usepackage[acceptedWithA]{tacl2018v2}

\usepackage{amssymb}
\usepackage{enumitem}
\usepackage{comment}

\usepackage{xspace,mfirstuc,tabulary}

\newif\iftaclinstructions
\taclinstructionsfalse 
\iftaclinstructions
\renewcommand{\confidential}{}
\renewcommand{\anonsubtext}{(No author info supplied here, for consistency with
TACL-submission anonymization requirements)}
\newcommand{\instr}
\fi

\iftaclpubformat 

\else

\fi

\usepackage{amsmath}
\usepackage{multirow}
\usepackage{booktabs}
\usepackage{graphicx}
\usepackage[normalem]{ulem}
\usepackage[scaled=.8]{beramono}

\usepackage[hang,flushmargin]{footmisc}
\newcommand\bart{\textsc{BART}}
\newcommand\tase{\textsc{TASE}}
\newcommand\tasedrop{\tase{}$_{\drop{}}$}
\newcommand\taseiirc{\tase{}$_{\iirc{}}$}
\newcommand\reader{\textsc{Reader}}
\newcommand\unifiedqa{\textsc{UnifiedQA}}
\newcommand\unifiedqadrop{\unifiedqa{}$_{\drop{}}$}
\newcommand\unifiedqahotpot{\unifiedqa{}$_{\textsc{HPQA}}$}
\newcommand\unifiedqaiirc{\unifiedqa{}$_{\iirc{}}$}

\newcommand\drop{\textsc{DROP}}
\newcommand\hotpotqa{\textsc{HotpotQA}}
\newcommand\iirc{\textsc{IIRC}}
\newcommand\breakdataset{\textsc{Break}}
\newcommand\squad{\textsc{SQuAD}}
\newcommand\boolq{\textsc{BoolQ}}

\newcommand\appendbool{\texttt{AppendBool}}
\newcommand\changelast{\texttt{ChangeLast}}
\newcommand\replacearith{\texttt{ReplaceArith}}
\newcommand\replacebool{\texttt{ReplaceBool}}
\newcommand\replacecomp{\texttt{ReplaceComp}}

\newcommand\prunestep{\texttt{PruneStep}}

\newcommand\constnumeric{\texttt{Numeric}}
\newcommand\constboolean{\texttt{Boolean}}
\newcommand\constgeq{\texttt{$\geq$}}
\newcommand\constleq{\texttt{$\leq$}}

\newcommand\cont{\textsc{CONT}}
\newcommand\contrand{\cont{}$_{\textsc{Rand}}$}
\newcommand\contval{\cont{}$_{\textsc{Val}}$}
\newcommand\const{\textsc{CONST}}
\newcommand\contconst{\cont{}$_{+\const{}}$}

\newcommand{\nl}[1]{\textit{``#1''}}
\newcommand\fone{F$_1$}
\newcommand{\meanstd}[2]{#1 $\pm$ #2}

\usepackage{xcolor, soul}
\definecolor{Background}{RGB}{217,245,203}
\sethlcolor{Background}

\title{\textit{Break, Perturb, Build}:
Automatic Perturbation of Reasoning Paths Through Question Decomposition}

\author{\bf Mor Geva}
\author{\bf Tomer Wolfson}
\author{\bf Jonathan Berant}

{
\makeatletter
\makeatother
\affil{School of Computer Science, Tel Aviv University}
\affil{Allen Institute for Artificial Intelligence}
}
\email{\normalsize \texttt{$\{$morgeva@mail,tomerwol@mail,joberant@cs$\}$.tau.ac.il}}

\date{}

\begin{document}
\maketitle

\begin{abstract}

Recent efforts to create challenge benchmarks that test the abilities of natural language understanding models have largely depended on human annotations. 
In this work, we introduce the \emph{``Break, Perturb, Build''} (BPB) framework for automatic reasoning-oriented perturbation of question-answer pairs. BPB represents a question by decomposing it into the reasoning steps that are required to answer it, 
symbolically perturbs the decomposition, and then generates new question-answer pairs. 
We demonstrate the effectiveness of BPB by creating evaluation sets for three reading comprehension (RC) benchmarks, generating thousands of high-quality examples without human intervention. We evaluate a range of RC models on our evaluation sets, which reveals large performance gaps on generated examples compared to the original data. Moreover, symbolic perturbations enable fine-grained analysis of the strengths and limitations of models. Last, augmenting the training data with examples generated by BPB helps close the performance gaps, without any drop on the original data distribution.

\end{abstract}

\begin{figure*}[t]\setlength{\belowcaptionskip}{-8pt}
    \centering
    \includegraphics[scale=0.47]{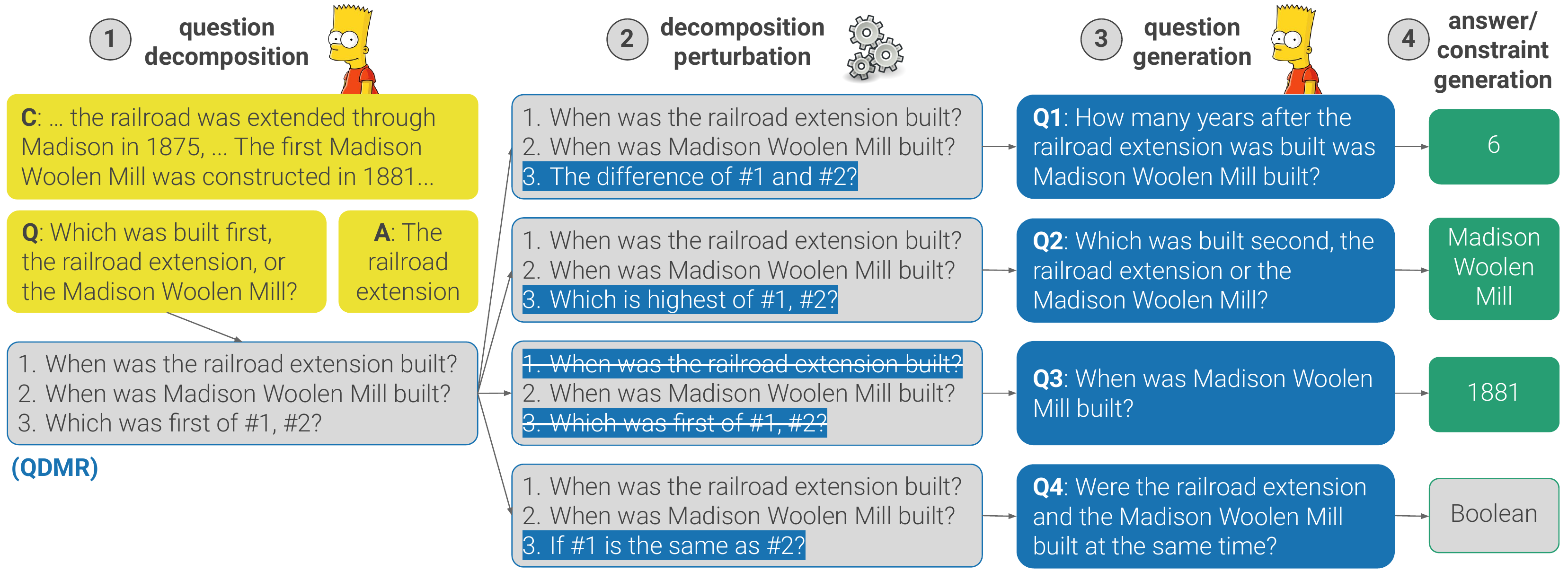}
    \vspace*{-2mm}
    \caption{An overview of BPB. Given a context (C), question (Q) and the answer (A) to the question, we generate new examples by 1) parsing the question into its QDMR decomposition, 2) applying semantic perturbations to the decomposition, 3) generating a question for each transformed decomposition, and 4) computing answers/constraints to the new questions.}
    \label{figure:overview}
\end{figure*}

\section{Introduction}
\label{sec:introduction}

Evaluating natural language understanding (NLU) systems has become a fickle enterprise. While models outperform humans on standard benchmarks, they perform poorly on a multitude of distribution shifts, (\citet{jia2017adversarial, naik2018stress, mccoy2019right}, \emph{inter alia}). 
To expose such gaps, recent work has proposed to evaluate models on \emph{contrast sets} \cite{gardner2020evaluating}, or \emph{counterfactually-augmented data} \cite{kaushik2020learning}, where minimal but meaningful perturbations are applied to test examples. However, since such examples are manually-written, collecting them is expensive, and procuring diverse perturbations is challenging \cite{joshi2021investigation}.

Recently, methods for automatic generation of contrast sets were proposed. However, current methods are restricted to shallow surface perturbations \cite{mille2021automatic, li2020linguistically}, specific reasoning skills \cite{asai2020logic}, or rely on expensive annotations \cite{bitton2021automatic}. 
Thus, automatic generation of examples that test high-level reasoning abilities of models and their robustness to fine semantic distinctions, remains an open challenge.

In this work, we propose the \emph{``Break, Perturb, Build''} (BPB) framework for automatic generation of reasoning-focused contrast sets for reading comprehension (RC).
Changing the high-level semantics of questions and generating question-answer pairs automatically is challenging. 
First, 
it requires extracting the reasoning path expressed in a question, in order to manipulate it.
Second, it requires the ability to generate grammatical and coherent questions.
In Fig.~\ref{figure:overview}, e.g., transforming Q, which involves \emph{number comparison}, into Q1, which requires \emph{subtraction}, leads to dramatic changes in surface form.
Third, it requires an automatic method for computing the answer to the perturbed question.

Our insight is that perturbing question semantics is possible when modifications are applied to a \emph{structured meaning representation}, rather than to the question itself.
Specifically, we represent questions with QDMR \cite{wolfson2020break}, a representation that decomposes a question into a sequence of reasoning steps, which are written in natural language and are easy to manipulate.
Relying on a structured representation lets us develop a pipeline for perturbing the reasoning path expressed in RC examples.

Our method (see Fig.~\ref{figure:overview}) has four steps. We (1) parse the question into its QDMR decomposition, (2) apply rule-based perturbations to the decomposition, (3) generate new questions from the perturbed decompositions, and (4) compute their answers. In cases where computing the answer is impossible, we compute constraints on the answer, which are also useful for evaluation. 
For example, for Q4 in Fig.~\ref{figure:overview}, even if we cannot extract the years of the described events, we know the answer type of the question (Boolean).
Notably, aside from answer generation, all steps depend on the question only, and can be applied to other modalities, such as visual or table question answering (QA).


Running BPB on the three RC datasets, \drop{} \cite{dua2019drop}, \hotpotqa{} \cite{yang2018hotpotqa}, and \iirc{} \cite{ferguson2020iirc}, yields thousands of semantically-rich examples, covering a majority of the original examples (63.5\%, 70.2\%, and 45.1\%, respectively).
Moreover, we validate examples using crowdworkers and find that $\geq$85\% of generated examples are correct. 

We demonstrate the utility of BPB for comprehensive and fine-grained evaluation of multiple RC models. First, we show that leading models, such as \unifiedqa{} \cite{khashabi2020unifiedqa} and \tase{} \cite{segal2020simple}, struggle on the generated contrast sets with a decrease of 13-36 \fone{} points and low consistency ($<$40).
Moreover, 
analyzing model performance per perturbation type and constraints, reveals the strengths and weaknesses of models on various reasoning types.
For instance, (a) models with specialized architectures are more brittle compared to general-purpose models trained on multiple datasets, (b) \tase{} fails to answer intermediate reasoning steps on \drop{}, (c) \unifiedqa{} fails completely on questions requiring numerical computations, and (d) models tend to do better when the numerical value of an answer is small.
Last, data augmentation with examples generated by BPB closes part of the performance gap, without any decrease on the original datasets.

In summary, we introduce a novel framework for automatic perturbation of complex reasoning questions, and demonstrate its efficacy for generating contrast sets and evaluating models.
We expect that imminent improvements in question generation, RC, and QDMR models will further widen the accuracy and applicability of our approach. The generated evaluation sets and codebase are publicly available at 
\url{https://github.com/mega002/qdmr-based-question-generation}.

\section{Background}
\label{sec:background}

Our goal, given a natural language question $q$, is to automatically alter its semantics, generating perturbed questions $\hat{q}$ for evaluating RC models.
This section provides background on the QDMR representation and the notion of \emph{contrast sets}.

\paragraph{Question Decomposition Meaning Representation (QDMR).} 
To manipulate question semantics, we rely on QDMR \cite{wolfson2020break}, a structured meaning representation for questions.
The QDMR decomposition $d=\text{QDMR}(q)$ is a sequence of reasoning steps $s_1, ..., s_{|d|}$ required to answer $q$. Each step $s_i$ in $d$ is an intermediate question, that is phrased in natural language and annotated with a logical operation $o_i$, such as \texttt{selection} (e.g. \nl{When was the Madison Woolen Mill built?}) or \texttt{comparison} (e.g. \nl{Which is highest of \#1, \#2?}).
Example QDMRs are shown in Fig.~\ref{figure:overview} (step 2).
QDMR paves a path towards controlling the reasoning path expressed in a question by changing, removing or adding steps (\S\ref{subsec:decomposition_transformation}).

\paragraph{Contrast sets.} 
\citet{gardner2020evaluating} defined the contrast set $\mathcal{C}(x)$ of an example $x$ with a label $y$ as a set of examples with  minimal perturbations to $x$ that typically affect $y$. Contrast sets evaluate whether a local decision boundary around an example is captured by a model.
In this work, given a question-context pair $x=\langle q,c \rangle$, we semantically perturb the question and generate examples $\hat{x}=\langle \hat{q},c \rangle \in \mathcal{C}(\langle q,c \rangle)$ that modify the original answer $a$ to $\hat{a}$.

\begin{table*}[t]\setlength{\belowcaptionskip}{-8pt}
\centering
\tiny
  \begin{tabular}{p{1.2cm}p{2.7cm}p{3.5cm}p{3.5cm}p{2.9cm}}
    Perturbation & Question & QDMR & Perturbed QDMR & Perturbed Question \\
    \toprule
    \texttt{Append Boolean step} & Kadeem Jack is a player in a league that started with how many teams? & (1) league that Kadeem Jack is a player in; (2) teams that \#1 started with; (3) number of \#2 & (1) league that Kadeem Jack is a player in; (2) teams that \#1 started with; (3) number of \#2; \hl{\bf(4) if \#3 is higher than 2} & \hl{If Kadeem Jack is a player in a league that started with more than two teams?}
  \\ \midrule
    \texttt{Change last step (to arithmetic)} & Which gallery was founded first, Hughes-Donahue Gallery or Art Euphoric? & (1) when was Hughes-Donahue Gallery founded; (2) when was Art Euphoric founded; \hl{\bf(3) which was first of \#1 , \#2} & (1) when was Hughes-Donahue Gallery founded; (2) when was Art Euphoric founded; \hl{\bf(3) the difference of \#1 and \#2} & \hl{How many years after Hughes-Donahue Gallery was founded was Art Euphoric founded?}  \\ \midrule
    \texttt{Change last step (to Boolean)} & How many years after Madrugada's final concert did Sunday Driver become popular? & (1) year of Madrugada's final concert; (2) year when Sunday Driver become popular; \hl{\bf(3) the difference of \#2 and \#1} & (1) year of Madrugada's final concert; (2) year when Sunday Driver become popular; \hl{\bf (3) if \#1 is the same as \#2} & \hl{Did Sunday Driver become popular in the same year as Madrugada's final concert?}  \\ \midrule
    \texttt{Replace arithmetic op.} & How many more native Hindi speakers are there compared to native Kannada speakers? & (1) native Hindi speakers; (2) native Kannada speakers; (3) number of \#1; (4) number of \#2; \hl{\bf(5) difference of \#3 and \#4} & (1) native Hindi speakers; (2) native Kannada speakers; (3) number of \#1; (4) number of \#2; \hl{\bf(5) sum of \#3 and \#4} & \hl{Of the native Hindi speakers and native Kannada speakers, how many are there in total?}  \\ \midrule
    \texttt{Replace Boolean op.} & Can Stenocereus and Pachypodium both include tree like plants? & (1) if Stenocereus include tree like plants; (2) if Pachypodium include treelike plants; \hl{\bf(3) if both \#1 and \#2 are true} & (1) if Stenocereus include tree like plants; (2) if Pachypodium include treelike plants; \hl{\bf(3) if both \#1 and \#2 are false} & \hl{Do neither Stenocereus nor Pachypodium include tree like plants?}  \\ \midrule
    \texttt{Replace comparison op.} & Which group is smaller for the county according to the census: people or households? & (1) size of the people group in the county according to the census; (2) size of households group in the county according to the census; \hl{\bf(3) which is smaller of \#1, \#2} & (1) size of the people group in the county according to the census; (2) size of households group in the county according to the census; \hl{\bf(3) which is highest of \#1, \#2} & \hl{According to the census, which group in the county from the county is larger: people or households?}  \\ \midrule
    \texttt{Prune step} & How many people comprised the total adult population of Cunter, excluding seniors? & (1) adult population of Cunter; \hl{\bf (2) \#1 excluding seniors;} (3) number of \#2 & (1) adult population of Cunter; (2) number of \#2 & \hl{How many adult population does Cunter have?}  \\

  \bottomrule
\end{tabular}
\caption{The full list of semantic perturbations in BPB. For each perturbation, we provide an example question and its decomposition. We highlight the altered decomposition steps, along with the generated question.}
\label{table:qdmr_transforms}
\end{table*}

\section{BPB: Automatically Generating Semantic Question Perturbations}
\label{sec:generation_pipeline}

We now describe the BPB framework.
Given an input $x = \langle q,c \rangle$ of question and context, and the answer $a$ to $q$ given $c$, we automatically map it to a set of new examples $\mathcal{C}(x)$ (Fig.~\ref{figure:overview}). Our approach uses models for question decomposition, question generation (QG), and RC.


\subsection{Question Decomposition}
\label{subsec:question_decomposition}

The first step (Fig.~\ref{figure:overview}, step 1) is to represent $q$ using a structured decomposition, $d = \text{QDMR}(q)$.
To this end, we train a text-to-text model that generates $d$ conditioned on $q$. Specifically, we fine-tune \bart{} \cite{lewis2020bart} on the \textit{high-level} subset of the \breakdataset{} dataset \cite{wolfson2020break}, which consists of 23.8K $\langle q,d\rangle$ pairs from three RC datasets, including \drop{} and \hotpotqa{}.\footnote{We fine-tune \bart{}-large for 10 epochs, using a learning rate of $3e^{-5}$ with polynomial decay and a batch size of 32.}
Our QDMR parser obtains a 77.3 SARI score on the development set, which is near state-of-the-art on the leaderboard.\footnote{\url{https://leaderboard.allenai.org/break_high_level/}}

\subsection{Decomposition Perturbation}
\label{subsec:decomposition_transformation}

A decomposition $d$ describes the reasoning steps necessary for answering $q$. By modifying $d$'s steps, we can control the semantics of the question. We define a ``library'' of rules for transforming $d \rightarrow \hat{d}$, and use it to generate questions $\hat{d} \rightarrow \hat{q}$. 

BPB provides a general method for creating a wide range of perturbations. In practice, though, deciding which rules to include is coupled with the reasoning abilities expected from our models. E.g., there is little point in testing a model on arithmetic operations if it had never seen such examples. Thus, we implement rules based on the reasoning skills required in current RC datasets \cite{yang2018hotpotqa, dua2019drop}.
As future benchmarks and models tackle a wider range of reasoning phenomena, one can expand the rule library.

Tab.~\ref{table:qdmr_transforms} provides examples for all QDMR perturbations, which we describe next:

\begin{itemize}[leftmargin=*,topsep=0pt,itemsep=0pt,parsep=0pt]
    \item \appendbool{}: When the question $q$ returns a numeric value, we transform its QDMR by appending a ``yes/no'' comparison step. The comparison is against the answer $a$ of question $q$. As shown in Tab.~\ref{table:qdmr_transforms}, the appended step compares the previous step result (\textit{``\#3''}) to a constant (\textit{``is higher than 2''}). \appendbool{} perturbations are generated for 5 comparison operators ($>, <, \leq, \geq, \neq$). For the compared values, we sample from a set, based on the answer $a$: $\{a+k , a-k, \frac{a}{k}, a\times k\}$ for $k \in \{1,2,3\}$. 
    \item \changelast{}: Changes the type of the last QDMR step. This perturbation is applied to steps involving operations over two referenced steps. Steps with type \{\texttt{arithmetic}, \texttt{comparison}\} have their type changed to either \{\texttt{arithmetic}, \texttt{Boolean}\}. 
    Tab.~\ref{table:qdmr_transforms} shows a comparison step changed to an \texttt{arithmetic} step, involving subtraction. Below it, an \texttt{arithmetic} step is changed to a yes/no question (\texttt{Boolean}).
    \item \replacearith{}:
    Given an \texttt{arithmetic} step, involving either subtraction or addition, we transform it by flipping its arithmetic operation.
    \item \replacebool{}: Given a \texttt{Boolean} step, verifying whether two statements are correct, we transform it to verify if neither are correct.
    \item \replacecomp{}: A \texttt{comparison} step compares two values and returns the highest or lowest. Given a \texttt{comparison} step, we flip its expression from ``highest'' to ``lowest'' and vice versa.
    \item \prunestep{}: We remove one of the QDMR steps. Following step pruning, we prune all other steps that are no longer referenced. We apply only a single \prunestep{} per $d$. Tab.~\ref{table:qdmr_transforms} displays $\hat{d}$ after its second step has been pruned.
\end{itemize}

\begin{table}[t]\setlength{\belowcaptionskip}{-8pt}
    \centering
    \scriptsize
    \begin{tabular}{p{2.8cm}p{3.9cm}}
         Original question & Augmented question \\ \toprule
         \textbf{How many} interceptions \textbf{did} Matt Hasselbeck throw? & \textbf{If} Matt Hasselbeck throw \textbf{less than 23} interceptions? (\hl{\appendbool{}}) \\ \hline
         \textbf{How many} touchdowns \textbf{were there} in the first quarter? & \textbf{If there were two} touchdowns in the first quarter? (\hl{\appendbool{}}) \\ \hline
         Are Giuseppe Verdi \textbf{and} Ambroise Thomas \textbf{both} Opera composers? & Are \textbf{neither} Giuseppe Verdi \textbf{nor} Ambroise Thomas Opera composers? (\hl{\replacebool{}}) \\ \hline
         Which singer is \textbf{younger}, Shirley Manson or Jim Kerr? & Which singer is \textbf{older}, Shirley Manson or Jim Kerr? (\hl{\replacecomp{}}) \\ \bottomrule
    \end{tabular}
    \caption{Example application of all textual patterns used to generate questions $q_{\textit{aug}}$ (perturbation type highlighted). Boldface indicates the pattern matched in $q$ and the modified part in $q_{\textit{aug}}$. Decompositions $d$ and $d_\textit{aug}$ omitted for brevity.
    }
    \label{table:question_gen_augment}
\end{table}

\subsection{Question Generation}
\label{subsec:question_generation}

At this point (Fig.~\ref{figure:overview}, step 3), we parsed $q$ to its decomposition $d$ and altered its steps to produce the perturbed decomposition $\hat{d}$. The new $\hat{d}$ expresses a different reasoning process compared to the original $q$. Next, we generate the \emph{perturbed question} $\hat{q}$ corresponding to $\hat{d}$. 
To this end, we train a QG model, generating questions conditioned on the input QDMR. Using the same $\langle q,d\rangle$ pairs used to train the QDMR parser (\S\ref{subsec:question_decomposition}), we train a separate \bart{} model for mapping $d \rightarrow q$.\footnote{We use the same hyperparameters as detailed in \S\ref{subsec:question_decomposition}, except the number of epochs, which was set to 15.}

An issue with our QG model is that the perturbed $\hat{d}$ may be outside the distribution the QG model was trained on, e.g., applying \appendbool{} on questions from \drop{} results in yes/no questions that do not occur in the original dataset. This can lead to low-quality questions $\hat{q}$.
To improve our QG model, we use simple heuristics to take $\langle q,d\rangle$ pairs from \breakdataset{} and generate additional pairs $\langle q_{\textit{aug}},d_{\textit{aug}
}\rangle$. Specifically, we define 4 textual patterns, associated with the  perturbations, \appendbool{}, \replacebool{} or \replacecomp{}. We automatically generate examples $\langle q_{\textit{aug}},d_{\textit{aug}} \rangle $ from $\langle q,d \rangle $ pairs that match a pattern. An example application of all patterns is in Tab.~\ref{table:question_gen_augment}. E.g. in \appendbool{}, the question $q_{\textit{aug}}$ is inferred with the pattern \textit{``how many ... did''}. In  \replacecomp{}, generating $q_{\textit{aug}}$ is done by identifying the superlative in $q$ and fetching its antonym.  

Overall, we generate 4,315 examples and train our QG model on the union of \breakdataset{} and the augmented data.
As QG models have been rapidly improving, we expect future QG models will be able to  generate high-quality questions for any decomposition without data augmentation.

\subsection{Answer Generation}
\label{subsec:answer_generation}
We converted the input question into a set of perturbed questions without using the answer or context. Therefore, this part of BPB can be applied to any question, regardless of the context modality. We now describe a RC-specific component for answer generation that uses the textual context.

To get complete RC examples, we must compute answers to the generated questions (Fig.~\ref{figure:overview}, step 4).
We take a two-step approach: 
For some questions, we can compute the answer automatically based on the type of applied perturbation. If this fails, we compute the answer by answering each step in the perturbed QDMR $\hat{d}$.

\paragraph{Answer generation methods.}
Let $\langle q,c,a \rangle$ be the original RC example and denote by $\hat{q}$ the generated question. We use the following per-perturbation rules to generate the new answer $\hat{a}$:
\begin{itemize}[leftmargin=*,topsep=0pt,itemsep=0pt,parsep=0pt]
    \item \appendbool{}: The transformed  $\hat{q}$ compares whether the answer $a$ and a numeric value $v$ satisfy a comparison condition. As the values of $a$ and $v$ are given (\S\ref{subsec:decomposition_transformation}),
    we can compute whether the answer is \nl{yes} or \nl{no} directly.
    
    \item \replacearith{}: This perturbation converts an answer that is the sum (difference) of numbers to an answer that is the difference (sum). 
    We can often identify the numbers by looking for numbers $x,y$ in the context $c$ such that $a = x\pm y$ and flipping the operation: $\hat{a} = |x\mp y|$. To avoid noise, we discard examples for which there is more than one pair of numbers that result in $a$, and cases where $a<10$, as the computation may involve explicit counting rather than an arithmetic computation.
    
    \item \replacebool{}: This perturbation turns a verification of whether two statements $x,y$ are true, to a verification of whether neither $x$ nor $y$ are true. Therefore, if $a$ is \nl{yes} (i.e. both $x,y$ are true), $\hat{a}$ must be \nl{no}.
    
    \item \replacecomp{}: This perturbation takes a comparison question $q$ that contains two candidate answers $x,y$, of which $x$ is the answer $a$.
    We parse $q$ with spaCy\footnote{\url{https://spacy.io/}.} and identify the two answer candidates $x,y$, and return the one that is not $a$.
\end{itemize}

\begin{figure}[t]\setlength{\belowcaptionskip}{-8pt}
    \centering
    \includegraphics[scale=0.48]{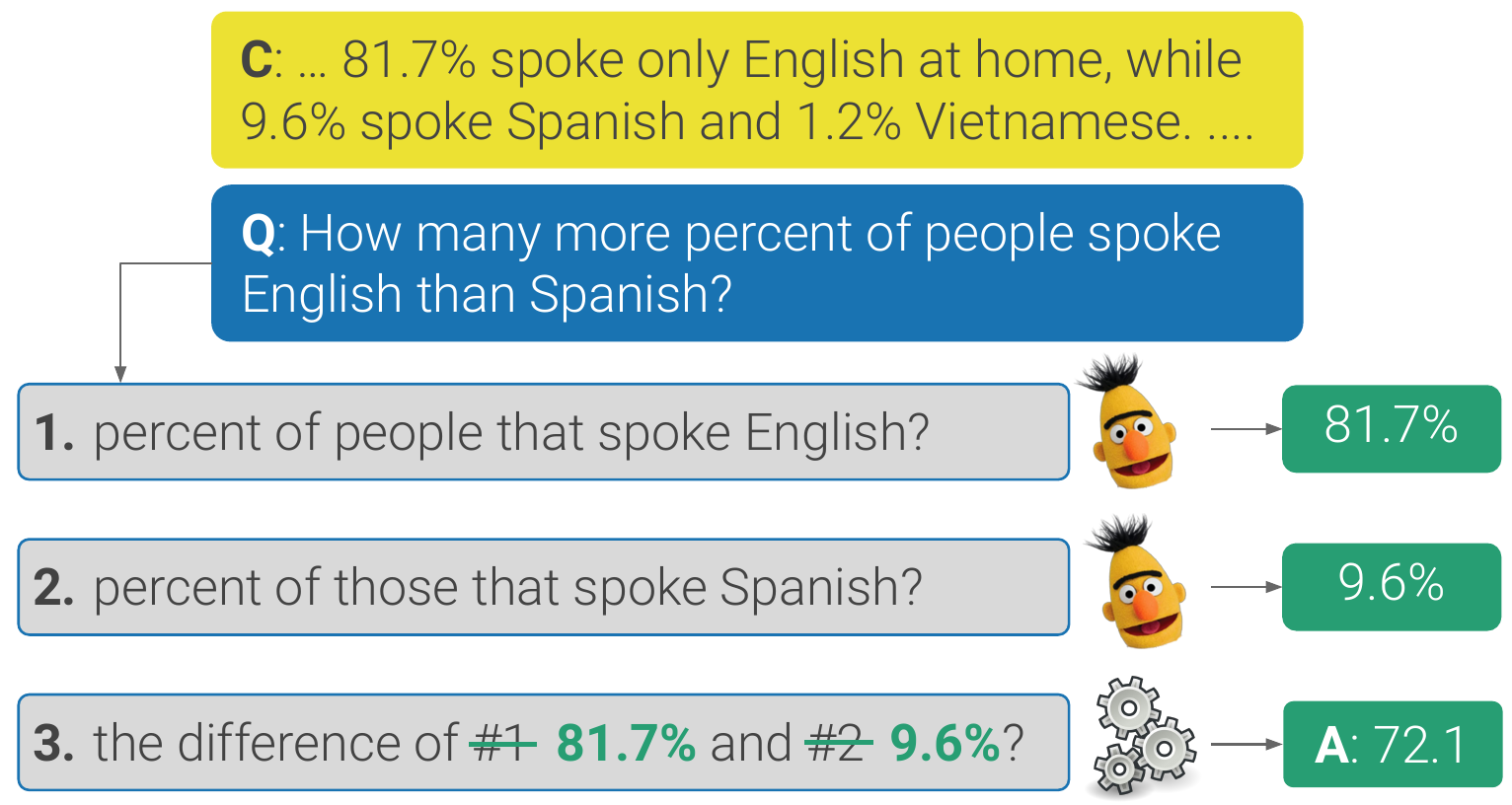}
    \vspace*{-2mm}
    \caption{Example execution of the QDMR evaluator.}
    \label{figure:qdmr_evaluator}
\end{figure}

\paragraph{QDMR evaluator.}
When our heuristics do not apply (e.g., arithmetic computations over more than two numbers, \prunestep{}, and \changelast{}), we use a RC model and the QDMR structure to directly \emph{evaluate} each step of $\hat{d}$ and compute $\hat{a}$.
Recall each QDMR step $s_i$ is annotated with a logical operation $o_i$ (\S\ref{sec:background}). 
To evaluate $\hat{d}$, we go over it step-by-step, and for each step either apply the RC model for operations that require querying the context (e.g. \texttt{selection}), or directly compute the output for numerical/set-based operations (e.g. \texttt{comparison}). 
The answer computed for each step is then used for replacing placeholders in subsequent steps. 
An example is provided in Fig.~\ref{figure:qdmr_evaluator}.

We discard the generated example when the RC model predicted an answer that does not match the expected argument type in a following step for which the answer is an argument (e.g. when a non-numerical span predicted by the RC model is used as an argument for an \texttt{arithmetic} operation), and when the generated answer has more than 8 words. Also, we discard operations that often produce noisy answers based on manual analysis (e.g. \texttt{project}  with a non-numeric answer).

For our QDMR evaluator, we fine-tune a \textsc{RoBERTa}-large model with a standard span-extraction output head on  \squad{} \cite{rajpurkar2016squad} and \boolq{} \cite{clark2019boolq}. \boolq{} is included to support yes/no answers.

\begin{table}[t]\setlength{\belowcaptionskip}{-8pt}
    \centering
    \footnotesize
    \begin{tabular}{p{3.2cm}|c|c|c}
         & \drop{} & \textsc{HPQA} & \iirc{}  \\ \toprule
         development set size & 9,536 & 7,405 & 1,301 \\ \hline
         \# of unique generated perturbations & 65,675 &  10,541 & 3,119 \\ \hline
         \# of generated examples & 61,231 & 8,488 & 2,450 \\ \hline
         \# of covered development examples & 6,053 & 5,199 & 587 \\ \hline
         \% of covered development examples & 63.5 & 70.2 & 45.1 \\ \midrule
         Avg. contrast set size & 11.1 & 2.6 & 5.2 \\ \hline
         Avg. \# of perturbations per example & 1.2 & 1  & 1 \\ \hline
         \% of answers generated by the QDMR evaluator & 5.8 & 61.8 & 22.5  \\
         \midrule
         \# of annotated contrast examples & 1,235 & 1,325 & 559 \\ \hline
         \% of valid annotated examples & 85 & 89 & 90.3 \\ \bottomrule
    \end{tabular}
    \caption{Generation and annotation statistics for the \drop{}, \hotpotqa{}, and \iirc{} datasets.}
    \label{table:generation_validation_stats}
\end{table}

\subsection{Answer Constraint Generation}
\label{subsec:answer_constraint_generation}

For some perturbations, even if we fail to generate an answer, it is still possible to derive constraints on the answer. Such constraints are valuable, as they indicate cases of model failure. Therefore, in addition to $\hat{a}$, we generate four types of \emph{answer constraints}: \constnumeric{}, \constboolean{}, \constgeq{}, \constleq{}.

When changing the last QDMR step to an arithmetic or Boolean operation (Tab.~\ref{table:qdmr_transforms}, rows 2-3), the new answer should be \constnumeric{} or \constboolean{}, respectively. An example for a \constboolean{} constraint is given in Q4 in Fig.~\ref{figure:overview}. 
When replacing an arithmetic operation (Tab.~\ref{table:qdmr_transforms}, row 4), if an answer that is the sum (difference) of two non-negative numbers is changed to the difference (sum) of these numbers, the new answer must not be greater (smaller) than the original answer. E.g., the answer to the question perturbed by \replacearith{} in Tab.~\ref{table:qdmr_transforms} (row 4) should satisfy the \constgeq{} constraint.

\begin{table}[t]\setlength{\belowcaptionskip}{-5pt}
    \centering
    \footnotesize
    \begin{tabular}{p{1.5cm}l|c|c|c}
         & & \drop{} & \textsc{HPQA} & \iirc{}  \\ \toprule
          & contrast & 56,205 & 2,754 & 1,884 \\
         \appendbool{} & annotated & 254 & 200 & 198 \\
         & \% valid & \textbf{97.2} & \textbf{98} & \textbf{98} \\ \hline
          & contrast & 85 & 408 & 43 \\
         \changelast{} & annotated & 69 & 200 & 43 \\
         & \% valid & \textbf{55.1} & \textbf{84.5} & \textbf{76.7} \\ \hline
          & contrast & 390 & - & 1 \\
         \replacearith{} & annotated & 191 & - & 1 \\
         & \% valid & \textbf{79.6} & - & 0 \\ \hline
          & contrast & - & 127 & 1 \\
         \replacebool{} & annotated & - & 127 & 1 \\
         & \% valid & - & \textbf{97.6} & 100 \\ \hline
          & contrast & 1,126 & 362 & 14 \\
         \replacecomp{} & annotated & 245 & 200 & 14 \\
         & \% valid & \textbf{90.2} & \textbf{88.5} & 71.4 \\ \hline
         & contrast & 3,425 & 3,777 & 507 \\
         \prunestep{} & annotated & 476 & 399 & 302 \\
         & \% valid & \textbf{82.4} & \textbf{85.8} & \textbf{88.4} \\ \hline
    \end{tabular}
    \caption{Per-perturbation statistics for generation and annotation of our datasets. Validation results are in bold for perturbations with at least 40 examples.}
    \label{table:generation_validation_stats_pre_transformation}
\end{table}

\section{Generated Evaluation Sets}
\label{sec:generated_data}

We run BPB on the RC datasets
\drop{} \cite{dua2019drop}, \hotpotqa{} \cite{yang2018hotpotqa}, and \iirc{} \cite{ferguson2020iirc}. Questions from the training sets of \drop{} and \hotpotqa{} are included in \breakdataset{}, and were used to train the decomposition and QG models. Results on \iirc{} show BPB's generalization to datasets for which we did not observe $\langle q, d \rangle$ pairs.
Statistics on the generated contrast and constraint sets are in Tab.~\ref{table:generation_validation_stats},~\ref{table:generation_validation_stats_pre_transformation} and~\ref{table:generated_constraint_stats}.

\paragraph{Contrast sets.}
Tab.~\ref{table:generation_validation_stats} shows that BPB successfully generates thousands of perturbations for each dataset. For the vast majority of perturbations, answer generation successfully produced a result -- for 61K out of 65K in \drop{}, 8.5K out of 10.5K in \hotpotqa{}, and 2.5K out of 3K in \iirc{}. Overall, 61K/8.5K examples were created from the development sets of \drop{}/ \hotpotqa{}, respectively, covering 63.5\%/70.2\% of the development set. For the held-out dataset \iirc{}, not used to train the QDMR parser and QG model, BPB created a contrast set of 2.5K examples, which covers almost half of the development set.

Tab.~\ref{table:generation_validation_stats_pre_transformation} shows the number of generated examples per perturbation. The distribution over perturbations is skewed, with some perturbations (\appendbool{}) 100x more frequent than others (\replacearith{}). This is since the original distribution over operations is not uniform and each perturbation operates on different decompositions (e.g., \appendbool{} can be applied to any question with a numeric answer, while \replacecomp{} operates on questions comparing two objects).

\paragraph{Constraint sets.}
Tab.~\ref{table:generated_constraint_stats} shows the number of generated answer constraints for each dataset.
The constraint set for \drop{} is the largest, consisting of 3.3K constraints, 8.9\% of which covering \drop{} examples for which we could not generate a contrast set. This is due to the examples with arithmetic operations, for which it is easier to generate constraints.
The constraint sets of \hotpotqa{} and \iirc{} contain yes/no questions, for which we use the \constboolean{} constraint.

\begin{table}[t]\setlength{\belowcaptionskip}{-8pt}
    \centering
    \footnotesize
    \begin{tabular}{p{3.cm}|c|c|c}
         & \drop{} & \textsc{HPQA} & \iirc{}  \\ \toprule
         \# of constraints & 3,323 & 550 & 56 \\ \hline
         \% of constraints that cover examples without a contrast set & 8.9 & 26 & 21.4 \\ \hline
         \% of covered development examples & 22.5 & 7.4 & 4 \\ \midrule
         \constnumeric{} & 2,398 & - & - \\ 
         \constboolean{} & - & 549 & 52 \\ 
         \constgeq{} & 825 & - & 1 \\ 
         \constleq{} & 100 & 1 & 3 \\ \hline
    \end{tabular}
    \caption{Generation of constraints statistics for the \drop{}, \hotpotqa{}, and \iirc{} datasets.}
    \label{table:generated_constraint_stats}
\end{table}

\begin{table*}[t]
    \centering
    \footnotesize
    \begin{tabular}{p{1.8cm}|c|c|c|c|c|c|c}
          & DEV & \contval{} & \contrand{} & \cont{} & \contval{} & \cont{} & \contconst{} \\
          & \fone{} & \fone{} & \fone{} & \fone{} & Cnst. & Cnst. & Cnst. \\\toprule
         \tasedrop{} & \meanstd{83.5}{0.1} & \meanstd{65.9}{1} & \meanstd{57.3}{0.6} & \meanstd{54.8}{0.4} & \meanstd{55.7}{1.1} & \meanstd{35.7}{0.5} & \meanstd{33.7}{0.3} \\
         \tasedrop{}+ & \meanstd{83.7}{1.1} & \meanstd{75.2}{0.5} & \meanstd{68}{1} & \meanstd{66.5}{0.5} & \meanstd{66.3}{0.4} & \meanstd{48.9}{0.6} & \meanstd{45}{0.4} \\ \midrule
         \taseiirc{} & \meanstd{69.9}{0.5} & \meanstd{45}{5} & \meanstd{41.2}{3.8} & \meanstd{33.7}{2.2} & \meanstd{23.7}{4.7} & \meanstd{24.3}{5.3} & \meanstd{24.3}{5.3} \\
         \taseiirc{}+ & \meanstd{68.8}{1.3} & \meanstd{81.1}{4.6} & \meanstd{78.2}{4.9} & \meanstd{72.4}{5.7} & \meanstd{50.4}{3.2} & \meanstd{48.2}{2.5} & \meanstd{48.2}{2.5} \\
         \toprule
    \end{tabular}
    \caption{Evaluation results of \tase{} on \drop{} and \iirc{}. For each dataset, we compare the model trained on the original and augmented (marked with +) training data.}
    \label{table:results_aggregated_tase}
\end{table*}

\begin{table*}[t]\setlength{\belowcaptionskip}{-8pt}
    \centering
    \footnotesize
    \begin{tabular}{p{2cm}|c|c|c|c|c|c|c}
          & DEV & \contval{} & \contrand{} & \cont{} & \contval{} & \cont{} & \contconst{} \\
           & \fone{} & \fone{} & \fone{} & \fone{} & Cnst. & Cnst. & Cnst. \\\toprule
         \reader{} & \meanstd{82.2}{0.2} & \meanstd{58.1}{0.1} & \meanstd{54.5}{0.7} & \meanstd{49.9}{0.4} & \meanstd{39.6}{0.6} & \meanstd{43.1}{0.1} & \meanstd{43}{0.1} \\
         \reader{}+ & \meanstd{82.7}{0.9} & \meanstd{89.1}{0.4} & \meanstd{86.6}{0.6} & \meanstd{81.9}{0.3} & \meanstd{65.6}{0.4} & \meanstd{56.4}{0.4} & \meanstd{56.3}{0.4} \\
         \toprule
    \end{tabular}
    \caption{Results of \reader{} on \hotpotqa{}, when trained on the original and augmented (marked with +) data.}
    \label{table:results_aggregated_reader}
\end{table*}

\begin{table*}[t]\setlength{\belowcaptionskip}{-8pt}
    \centering
    \footnotesize
    \begin{tabular}{p{2.2cm}|c|c|c|c|c|c|c}
          & DEV & \contval{} & \contrand{} & \cont{} & \contval{} & \cont{} & \contconst{} \\
           & \fone{} & \fone{} & \fone{} & \fone{} & Cnst. & Cnst. & Cnst. \\\toprule
         \unifiedqa{} & 28.2 & 38.1 & 35.1 & 34.9 & 5.3 & 4.4 & 2.2 \\
         \unifiedqadrop{} & \meanstd{33.9}{0.9} & \meanstd{28.4}{0.8} & \meanstd{26.9}{0.5} & \meanstd{8.1}{3.8} & \meanstd{12.2}{1.6} & \meanstd{5.1}{0.7} & \meanstd{4.4}{0.5} \\
         \unifiedqadrop{}+ & \meanstd{32.9}{1.2} & \meanstd{37.9}{1.4} & \meanstd{35.9}{2.5} & \meanstd{10.5}{4.4} & \meanstd{16.9}{0.2} & \meanstd{9.6}{0.2} & \meanstd{8}{0.5} \\ \midrule
         \unifiedqa{} & 68.7 & 68.2 & 52.9 & 65.2 & 29.8 & 38.4 & 37.6 \\
         \unifiedqahotpot{} & \meanstd{74.7}{0.2} & \meanstd{60.3}{0.8} & \meanstd{58.7}{0.9} & \meanstd{61.9}{0.7} & \meanstd{35.6}{1.1} & \meanstd{40.2}{0.1} & \meanstd{39.9}{0.1} \\
         \unifiedqahotpot{}+ & \meanstd{74.1}{0.2} & \meanstd{60.3}{1.9} & \meanstd{59.2}{1.5} & \meanstd{62.3}{2.3} & \meanstd{36.3}{0.7} & \meanstd{41.6}{0.3} & \meanstd{41.3}{0.4} \\ \midrule
         \unifiedqa{} & 44.5 & 61.1 & 57.2 & 36.5 & 21.6 & 28.1 & 28.1 \\
         \unifiedqaiirc{} & \meanstd{50.2}{0.7} & \meanstd{45.1}{2.1} & \meanstd{42.5}{2.3} & \meanstd{20.4}{2.9} & \meanstd{24.9}{1.2} & \meanstd{28.6}{0.8} & \meanstd{28.5}{0.8} \\
         \unifiedqaiirc{}+ & \meanstd{51.7}{0.9} & \meanstd{62.9}{2.9} & \meanstd{54.5}{3.9} & \meanstd{40.8}{5.4} & \meanstd{30.2}{2.7} & \meanstd{32.1}{1.9} & \meanstd{32.1}{1.9} \\
         \toprule
    \end{tabular}
    \caption{Evaluation results of \unifiedqa{} on \drop{}, \hotpotqa{}, and \iirc{}. We compare \unifiedqa{} without fine-tuning, and after fine-tuning on the original training data and on the augmented training data (marked with +).}
    \label{table:results_aggregated_unifiedqa}
\end{table*}

\paragraph{Estimating Example Quality}
\label{subsec:validated_contrast_sets}
To analyze the quality of generated examples, we sampled 200-500 examples from each perturbation and dataset (unless fewer than 200 examples were generated) and let crowdworkers validate their correctness. We qualify 5 workers, and establish a feedback protocol where we review work and send feedback after every annotation batch \cite{nangia2021ingredients}. Each generated example was validated by three workers, and is considered valid if approved by the majority. Overall, we observe a Fleiss Kappa \cite{fleiss1971measuring} of 0.71, indicating substantial annotator agreement \cite{landis1977measurement}.

Results are in Tab.~\ref{table:generation_validation_stats}, \ref{table:generation_validation_stats_pre_transformation}. The vast majority of generated examples ($\geq$85\%) were marked as valid, showing that BPB produces high-quality examples.
Moreover (Tab.~\ref{table:generation_validation_stats_pre_transformation}), we see variance across perturbations, where some perturbations reach >95\% valid examples (\appendbool{}, \replacebool{}), while others (\changelast{}) have lower validity. Thus, overall quality can be controlled by choosing specific perturbations.

Manual validation of generated contrast sets is cheaper than authoring contrast sets from scratch: The median validation time per example is 31 seconds, roughly an order of magnitude faster than reported in \newcite{gardner2020evaluating}.
Thus, when a very clean evaluation set is needed, BPB can dramatically reduce the cost of manual annotation.

\paragraph{Error Analysis of the QDMR Parser}
To study the impact of errors by the QDMR parser on the quality of generated examples, we (the authors) took the examples annotated by crowdworkers, and analyzed the generated QDMRs for 60 examples per perturbation from each dataset: 30 that were marked as valid by crowdworkers, and 30 that were marked as invalid. 
Specifically, for each example, we checked whether the generated QDMR faithfully expresses the reasoning path required to answer the question, and compared the quality of QDMRs of valid and invalid examples.

For the examples that were marked as valid, we observed that the accuracy of QDMR structures is high: 89.5\%, 92.7\%, and 91.1\% for \drop{}, \hotpotqa{}, and \iirc{}, respectively. This implies that, overall, our QDMR parser generated faithful and accurate representations for the input questions. Moreover, for examples marked as invalid, the QDMR parser accuracy was lower but still relatively high, with 82.0\%, 82.9\%, and 75.5\% valid QDMRs for \drop{}, \hotpotqa{}, and \iirc{}, respectively. This suggests that the impact of errors made by the QDMR parser on generated examples is moderate.


\section{Experimental Setting}
\label{sec:experimental_setting}

We use the generated contrast and constraints sets to evaluate the performance of strong RC models. 

\subsection{Models}
\label{subsec:models}

To evaluate our approach, we examine a suite of models that perform well on current RC benchmarks, and that are diverse it terms of their architecture and the reasoning skills they address:
\begin{itemize}[leftmargin=*,topsep=0pt,itemsep=0pt,parsep=0pt]
    \item \tase{} \cite{segal2020simple}: A \textsc{RoBERTa} model \cite{liu2019roberta} with 4 specialized output heads for (a) tag-based multi-span extraction, (b) single-span extraction, (c) signed number combinations, and (d) counting (until 9). \tase{} obtains near state-of-the-art performance when fine-tuned on \drop{}.
    \item \unifiedqa{} \cite{khashabi2020unifiedqa}: A text-to-text \textsc{T5} model \cite{raffel2020exploring} that was fine-tuned on multiple QA datasets with different answer formats (e.g. yes/no, span, etc.). \unifiedqa{} has demonstrated high performance on a wide range of QA benchmarks.
    \item \reader{} \cite{asai2020learning}: A \textsc{BERT}-based model \cite{devlin2018bert} for RC with two output heads for answer classification to \texttt{yes}/\texttt{no}/\texttt{span}/\texttt{no-answer}, and span extraction.
\end{itemize}

We fine-tune two \tase{} models, one on \drop{} and another on \iirc{}, which also requires numerical reasoning. \reader{} is fine-tuned on \hotpotqa{}, while separate \unifiedqa{} models are fine-tuned on each of the three datasets.
In addition, we evaluate \unifiedqa{} without fine-tuning, to analyze its generalization to unseen QA distributions. We denote by \unifiedqa{} the model without fine-tuning, and by \unifiedqa{}$_\textsc{X}$ the \unifiedqa{} model fine-tuned on dataset \textsc{X}.

We consider a ``pure'' RC setting, where only the context necessary for answering is given as input. For \hotpotqa{}, we feed the model with the two gold paragraphs (without distractors), and for \iirc{} we concatenate the input paragraph with the gold evidence pieces from other paragraphs.

Overall, we study 6 model-dataset combinations, with 2 models per dataset. For each model, we perform a hyperparameter search and train 3-4 instances with different random seeds, using the best configuration on the development set.

\subsection{Evaluation}
\label{subsec:evaluation}

We evaluate each model in multiple settings: (a) the original development set; (b) the generated contrast set, denoted by \cont{}; (c) the subset of \cont{} marked as valid by crowdworkers, denoted by \contval{}.  
Notably, \cont{} and \contval{} have a different distribution over perturbations. To account for this discrepancy, we also evaluate models on a sample from \cont{}, denoted by \contrand{}, where sampling is according to the perturbation distribution in \contval{}. Last, to assess the utility of constraint sets, we enrich the contrast set of each example with its corresponding constraints, denoted by \contconst{}.

Performance is measured using the standard \fone{} metric. In addition, we measure \textit{consistency} \cite{gardner2020evaluating}, that is, the fraction of examples such that the model predicted the correct answer to the original example as well as to all examples generated for this example. A prediction is considered correct if the \fone{} score, with respect to the gold answer, is $\geq 0.8$. Formally, for a set of evaluation examples $\mathcal{S}=\{\langle q_i,c_i,a_i \rangle\}_{i=1}^{|\mathcal{S}|}$:
$$ \textit{consistency}(\mathcal{S}) = \frac{1}{|\mathcal{S}|} \sum_{x \in \mathcal{S}} g(\mathcal{C}(x)) $$
$$ g(\mathcal{X}) =
\begin{cases}
1, & \text{if } \forall \langle \hat{x},\hat{a} \rangle \in \mathcal{X}: \text{F}_1(y(\hat{x}), \hat{a}) \geq 0.8 \\
0, & \text{otherwise}
\end{cases} $$
where $\mathcal{C}(x)$ is the generated contrast set for example $x$ (which includes $x$),\footnote{With a slight abuse of notation, we overload the definition of $\mathcal{C}(x)$ from \S\ref{sec:background}, such that members of $\mathcal{C}(x)$ include not just the queston and context, but also an answer.} and $y(\hat{x})$ is the model's prediction for example $\hat{x}$.
Constraint satisfaction is measured using a binary 0-1 score.

Since yes/no questions do not exist in \drop{}, we do not evaluate \tasedrop{} on \appendbool{} examples, which have yes/no answers, as we cannot expect the model to answer those correctly.

\subsection{Results}
\label{subsec:aggregated_results}

Results are presented separately for each model, in Tab.~\ref{table:results_aggregated_tase},~\ref{table:results_aggregated_reader} and~\ref{table:results_aggregated_unifiedqa}.
Comparing performance on the development sets (DEV \fone{}) to the corresponding contrast sets (\cont{} \fone{}), we see a substantial decrease in performance on the generated contrast sets, across all datasets (e.g. 83.5 $\rightarrow$ 54.8 for \tasedrop{}, 82.2 $\rightarrow$ 49.9 for \reader{}, and 50.2 $\rightarrow$ 20.4 for \unifiedqaiirc{}). Moreover, model consistency (\cont{} Cnst.) is considerably lower than the development scores (DEV \fone{}), for example, \taseiirc{} obtains 69.9 \fone{} score but only 24.3 consistency. This suggests that, overall, the models 
do not generalize to pertrubations in the reasoning path expressed in the original question.

Comparing the results on the contrast sets and their validated subsets (\cont{} vs. \contval{}), performance on \contval{} is better than on \cont{} (e.g., 58.1 versus 49.9 for \reader{}). These gaps are due to (a) the distribution mismatch between the two sets, and (b) bad example generation. To isolate the effect of bad example generation, we can compare \contval{} to \contrand{}, which have the same distribution over perturbations, but \contrand{} is not validated by humans. We see that the performance of \contval{} is typically $\leq$10\% higher than \contrand{} (e.g., 58.1 vs. 54.5 for \reader{}). Given that performance on the original development set is dramatically higher, it seems we can currently use automatically-generated contrast sets (without verification) to evaluate robustness to reasoning perturbations.


Last, adding constraints to the generated contrast sets (\cont{} vs. \contconst{}) often leads to a decrease in model consistency, most notably on \drop{}, where there are arithmetic constraints and not only answer type constraints.
For instance, consistency drops from 35.7 to 33.7 for \tase{}, and from 5.1 to 4.4 for \unifiedqadrop{}.
This shows that the generated constraints expose additional flaws in current models.


\subsection{Data Augmentation}
\label{subsec:data_augmentation}

Results in \S\ref{subsec:aggregated_results} reveal clear performance gaps in current QA models.
A natural solution is to augment the training data with examples from the contrast set distribution, which can be done effortlessly, since BPB is fully automatic.

We run BPB on the training sets of \drop{}, \hotpotqa{}, and \iirc{}. As BPB generates many examples, it can shift the original training distribution dramatically. Thus, we limit the number of examples generated by each perturbation by a threshold $\tau$. Specifically, for a training set $\mathcal{S}$ with $|\mathcal{S}|=n$ examples, we augment $\mathcal{S}$ with $\tau * n$ randomly generated examples from each perturbation (if less than $\tau * n$ examples were generate we add all of them). We experiment with three values $\tau \in \{0.03, 0.05, 0.1\}$, and choose the trained model with the best \fone{} on the contrast set. 

Augmentation results are shown in Tab.~\ref{table:results_aggregated_tase}-\ref{table:results_aggregated_unifiedqa}. Consistency (\cont{} and \contval{}) improves dramatically with only a small change in the model's DEV performance, across all models. We observe an increase in consistency of 13 points for \tasedrop{}, 24 for \taseiirc{}, 13 for \reader{}, and 1-4 points for the \unifiedqa{} models. Interestingly, augmentation is less helpful for \unifiedqa{} than for \tase{} and \reader{}. 
We conjecture that this is since \unifiedqa{} was trained on examples from multiple QA datasets and is thus less affected by the augmented data.

Improvement on test examples sampled from the augmented training distribution is expected. To test whether augmented data improves robustness on other distributions, we evaluate \tase{}+ and \unifiedqadrop{}+  on the \drop{} contrast set manually collected by \citet{gardner2020evaluating}. 
We find that training on the augmented training set does not lead to a significant change on the manually collected contrast set (\fone{}of 60.4 $\rightarrow$ 61.1 for \tase{}, and 30 $\rightarrow$ 29.6 for \unifiedqadrop{}).
This agrees with findings that data augmentation w.r.t a phenomenon may not improve generalization to other out-of-distribution examples \cite{kaushik2021efficacy, joshi2021investigation}.



\section{Performance Analysis}
\label{sec:performance_analysis}

\begin{figure}[t]\setlength{\belowcaptionskip}{-5pt}
    \centering
    \includegraphics[width=\columnwidth]{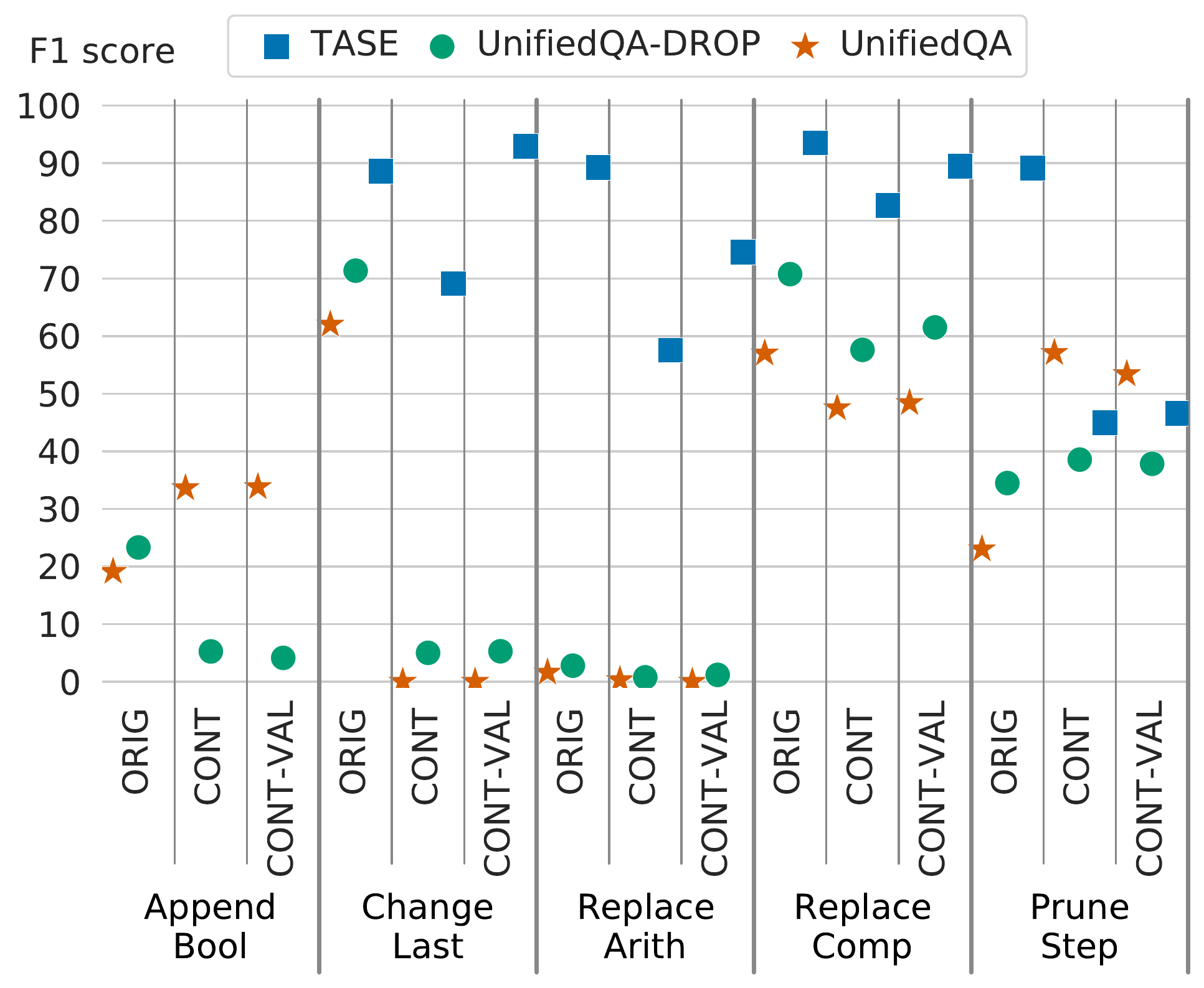}
    \caption{Performance on \drop{} per perturbation: on the generated contrast set (\cont{}), on the examples from which \cont{} was generated (\textsc{ORIG}), and on the validated subset of \cont{} (\contval{}).}
    \label{figure:performance_breakdown_drop}
\end{figure}
\begin{figure}[t]\setlength{\belowcaptionskip}{-5pt}
    \centering
    \includegraphics[width=\columnwidth]{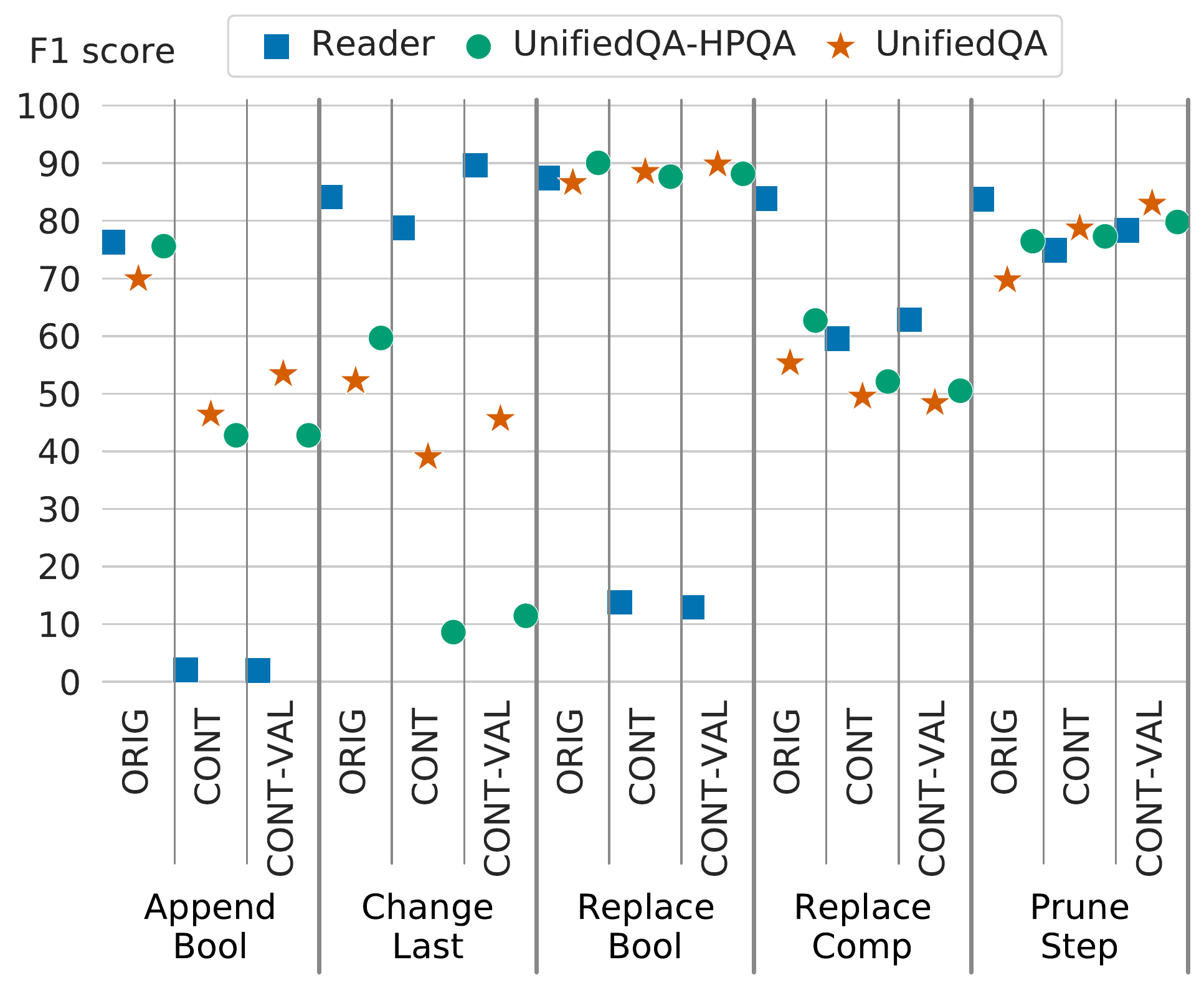}
    \caption{Performance on \hotpotqa{} per perturbation: on the generated contrast set (\cont{}), on the examples from which \cont{} was generated (\textsc{ORIG}), and the validated subset of \cont{} (\contval{}).}
    \label{figure:performance_breakdown_hotpotqa}
\end{figure}
\begin{figure}[t]\setlength{\belowcaptionskip}{-5pt}
    \centering
    \includegraphics[scale=0.4]{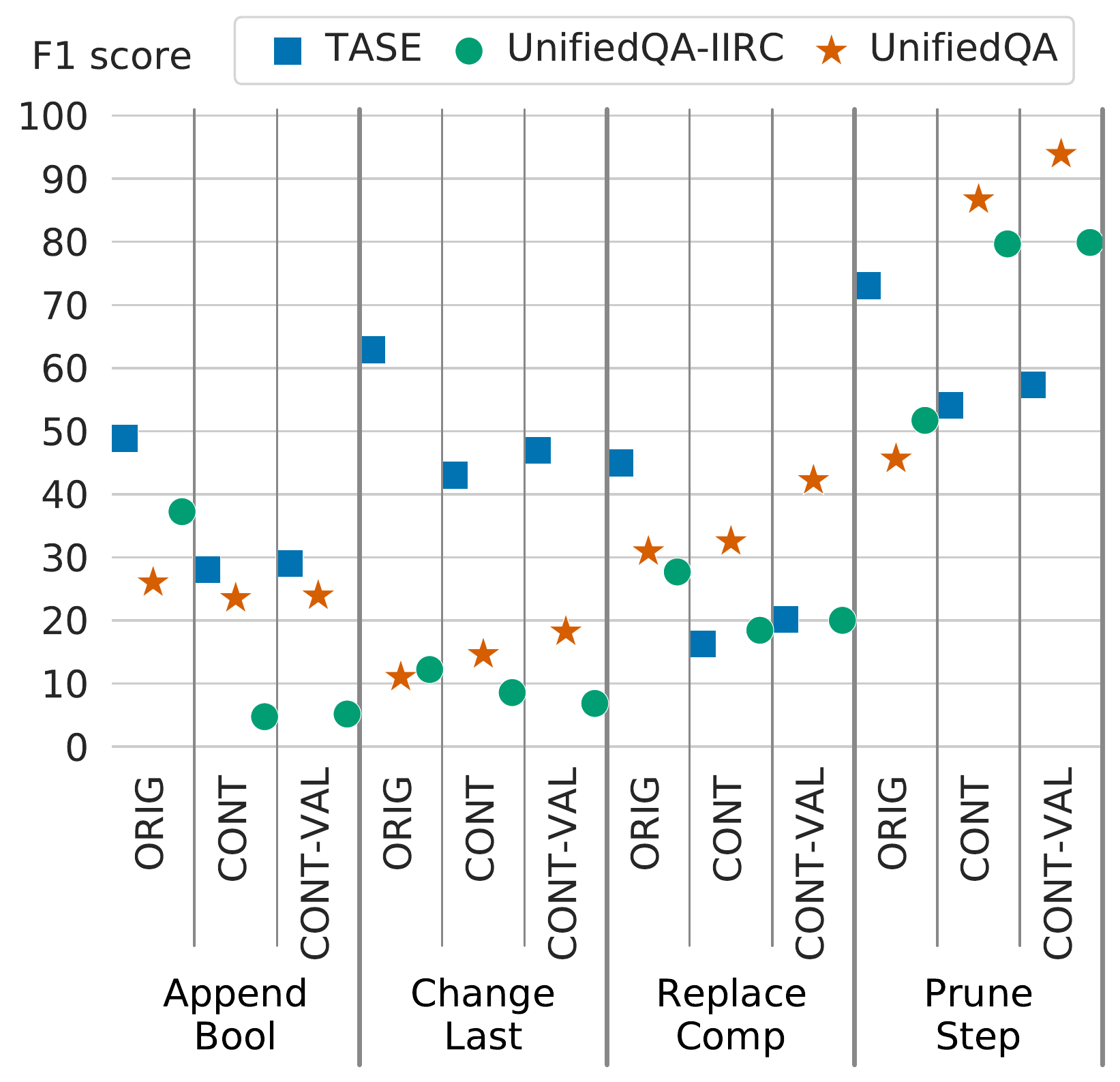}
    \caption{Performance on \iirc{} per perturbation: on the generated contrast set (\cont{}), on the examples from which \cont{} was generated (\textsc{ORIG}), and the validated subset of \cont{} (\contval{}).}
    \label{figure:performance_breakdown_iirc}
\end{figure}

\paragraph{Analysis across perturbations.}
We compare model performance on the original (ORIG) and generated examples (\cont{} and \contval{}) across perturbations (Fig.~\ref{figure:performance_breakdown_drop},~\ref{figure:performance_breakdown_hotpotqa},~\ref{figure:performance_breakdown_iirc}). 
Starting from models with specialized architectures (\tase{} and \reader{}), except for \changelast{} (discussed later), models' performance decreases on all perturbations. Specifically, \tase{} (Fig.~\ref{figure:performance_breakdown_drop},~\ref{figure:performance_breakdown_iirc}) demonstrates brittleness to changes in comparison questions (10-30 \fone{} decrease on \replacecomp{}) and arithmetic computations ($\sim$30 \fone{} decrease on \replacearith{}).  The biggest decrease of almost 50 points is on examples generated by \prunestep{} from \drop{} (Fig.~\ref{figure:performance_breakdown_drop}), showing that the model struggles to answer intermediate reasoning steps. 

\reader{} (Fig.~\ref{figure:performance_breakdown_hotpotqa}) shows similar trends to \tase{}, with a dramatic performance decrease of 80-90 points on yes/no questions created by \appendbool{} and \replacebool{}. 
Interestingly, \reader{} obtains high performance on \prunestep{} examples, as opposed to \tasedrop{} (Fig.~\ref{figure:performance_breakdown_drop}), which has a similar span extraction head that is required for these examples. This is possibly due to the ``train-easy'' subset of \hotpotqa{}, which includes single-step selection questions.

Moving to the general-purpose \unifiedqa{} models, they perform on \prunestep{} at least as well the original examples, showing their ability to answer simple selection questions. They also demonstrate robustness on \replacebool{}. Yet, they struggle on numeric comparison questions or arithmetic calculations: $\sim$65 points decrease on \changelast{} on \drop{} (Fig.~\ref{figure:performance_breakdown_drop}), 10-30 \fone{} decrease on \replacecomp{} and \appendbool{} (Fig.~\ref{figure:performance_breakdown_drop},~\ref{figure:performance_breakdown_hotpotqa},~\ref{figure:performance_breakdown_iirc}), and almost 0 \fone{} on \replacearith{} (Fig.~\ref{figure:performance_breakdown_drop}).

\paragraph{Performance on \cont{} and \contval{}.}
Results on \contval{} are generally higher than \cont{} due to the noise in example generation. However, whenever results on ORIG are higher than \cont{}, they are also higher than \contval{}, showing that the general trend can be inferred from \cont{}, due to the large performance gap between ORIG and \cont{}.
An exception is \changelast{} in \drop{} and \hotpotqa{}, where performance on \cont{} is lower than ORIG, but on \contval{} is higher. This is probably due to the noise in generation, especially for \drop{}, where example validity is at 55.1\% (see
Tab.~\ref{table:generation_validation_stats_pre_transformation}).

\paragraph{Evaluation on answer constraints}

Evaluating whether the model satisfies answer constraints can help assess the model's skills. To this end, we measure the fraction of answer constraints satisfied by the predictions of each model (we consider only constraints with more than 50 examples).

Models typically predict the correct answer type; \tasedrop{} and \unifiedqa{} predict a number for $\geq86$\% of the generated numeric questions, and \reader{} and \taseiirc{} successfully predict a yes/no answer in $\geq92$\% of the cases. However, fine-tuning \unifiedqa{} on \hotpotqa{} and \iirc{} reduces constraint satisfaction (94.7 $\rightarrow$ 76.3 for \unifiedqahotpot, 65.4 $\rightarrow$ 38.9 for \unifiedqaiirc{}), possibly since yes/no questions comprise less than 10\% of the examples \cite{yang2018hotpotqa, ferguson2020iirc}.
In addition, results on \drop{} for the constraint `\constgeq{}' are considerably lower than for `\constleq{}' for \unifiedqa{} ($83 \rightarrow 67.4$) and \unifiedqadrop{} ($81.8 \rightarrow 65.9$), indicating a bias towards predicting small numbers. 

\section{Related Work}
\label{sec:related_work}

The evaluation crisis in NLU has led to wide interest in challenge sets that evaluate the robustness of models to input perturbations. However, most past approaches \cite{ribeiro2020beyond, gardner2020evaluating, khashabi2020bang, kaushik2020learning} involve a human-in-the-loop and are thus costly.

Recently, more and more work considered using meaning representations of language to automatically generate evaluation sets. Past work used an ERG grammar \cite{li2020linguistically} and AMR \cite{rakshit2021asq} to generate relatively shallow perturbations. In parallel to this work, \newcite{ross2021tailor} used control codes over SRL to generate more semantic perturbations to declarative sentences. We generate perturbations at the level of the \emph{underlying reasoning process}, in the context of QA. Last, \newcite{bitton2021automatic} used scene graphs to generate examples for visual QA. However, they assumed the existence of gold scene graph at the input. Overall, this body of work represents an exciting new research program, where structured representations are leveraged to test and improve the blind spots of pre-trained language models.

More broadly, interest in automatic creation of evaluation sets that test out-of-distribution generalization has skyrocketed, whether using heuristics \cite{asai2020logic,wu2021polyjuice}, data splits \cite{finegan-dollak2018improving, keysers2020measuring}, adversarial methods \cite{alzantot2018generating}, or an aggregation of the above \cite{mille2021automatic,goel2021robustness}.

Last, QDMR-to-question generation is broadly related to work on text generation from structured data \cite{nan2021dart, novikova2017e2e,shu2021logic}, and to passage-to-question generation methods \cite{du2017learning,wang2020pathqg,duan2017question} that, in contrast to our work, focused on simple questions not requiring reasoning.



\section{Discussion}
\label{sec:conclusion}

We propose the BPB framework for generating high-quality reasoning-focused question perturbations, and demonstrate its utility for constructing contrast sets and evaluating RC models. 

While we focus on RC, our method for perturbing questions is independent of the context modality. Thus, porting our approach to other modalities only requires a method for computing the answer to perturbed questions. Moreover, BPB provides a general-purpose mechanism for question generation, which can be used outside QA as well.

We provide a library of perturbations that is a function of the current abilities of RC models. As future RC models, QDMR parsers, and QG models improve, we can expand this library to support additional semantic phenomena. 

Last, we showed that constraint sets are useful for evaluation. Future work can use constraints as a supervision signal, similar to \newcite{dua2021learning}, who leveraged dependencies between training examples to enhance model performance. 

\paragraph{Limitations}
BPB represents questions with QDMR, which is geared towards representing complex factoid questions that involve multiple reasoning steps. 
Thus, BPB cannot be used when questions involve a single step, e.g., one cannot use BPB to perturb \nl{Where was Barack Obama born?}. 
Inherently, the effectiveness of our pipeline approach depends on the performance of its modules -- the QDMR parser, the QG model, and the single-hop RC model used for QDMR evaluation. However, our results suggest that current models already yield high-quality examples, and model performance is expected to improve over time.

\section*{Acknowledgement}
We thank Yuxiang Wu, Itay Levy and Inbar Oren for the helpful feedback and suggestions. This research was supported in part by The Yandex Initiative for Machine Learning, and The European Research Council (ERC) under the European Union Horizons 2020 research and innovation programme (grant ERC DELPHI 802800). This work was completed in partial fulfillment for the Ph.D degree of Mor Geva.

\bibliography{all}
\bibliographystyle{acl_natbib}

\end{document}